\PassOptionsToPackage{table}{xcolor}  
\PassOptionsToPackage{capitalize}{cleveref}  
\PassOptionsToPackage{headheight=13.5mm,headsep=3mm}{geometry}
\documentclass{bytedance_seed}

\usepackage{amsmath,amssymb}
\usepackage{array}
\usepackage{xspace}

\AtBeginEnvironment{tabular}{\renewcommand{\textbf}[1]{{\bfseries #1}}}


\definecolor{oursrow}{HTML}{EEEAFB}
\definecolor{tabgray}{HTML}{52514E}
\definecolor{notespurple}{HTML}{6A4FD0}
\definecolor{videoorange}{HTML}{EB6834}

\newtcolorbox{findingbox}{colback=oursrow,colframe=notespurple!70,boxrule=0.5pt,arc=1.5pt,
  left=3pt,right=3pt,top=2pt,bottom=2pt}
\newcounter{finding}
\newcommand{\finding}[2]{\refstepcounter{finding}%
  \begin{findingbox}\small\textbf{Finding~\thefinding. #1} #2\end{findingbox}}

\input{numbers}

\newcommand{\method}{LookAway\xspace}

\newcommand{\x}{$\times$}

\newenvironment{sidecaption}[2]{%
  \begin{minipage}[c]{#1\textwidth}\includegraphics[width=\linewidth]{#2}\end{minipage}\hfill
  \begin{minipage}[c]{\dimexpr0.96\textwidth-#1\textwidth\relax}\captionsetup{skip=0pt}}
  {\end{minipage}}

\makeatletter
\DeclareRobustCommand\onedot{\futurelet\@let@token\@onedot}
\def\@onedot{\ifx\@let@token.\else.\null\fi\xspace}

\makeatother

\fancypagestyle{firststyle}{%
  \fancyhead[R]{}%
  \fancyhead[L]{%
    \includegraphics[height=12mm]{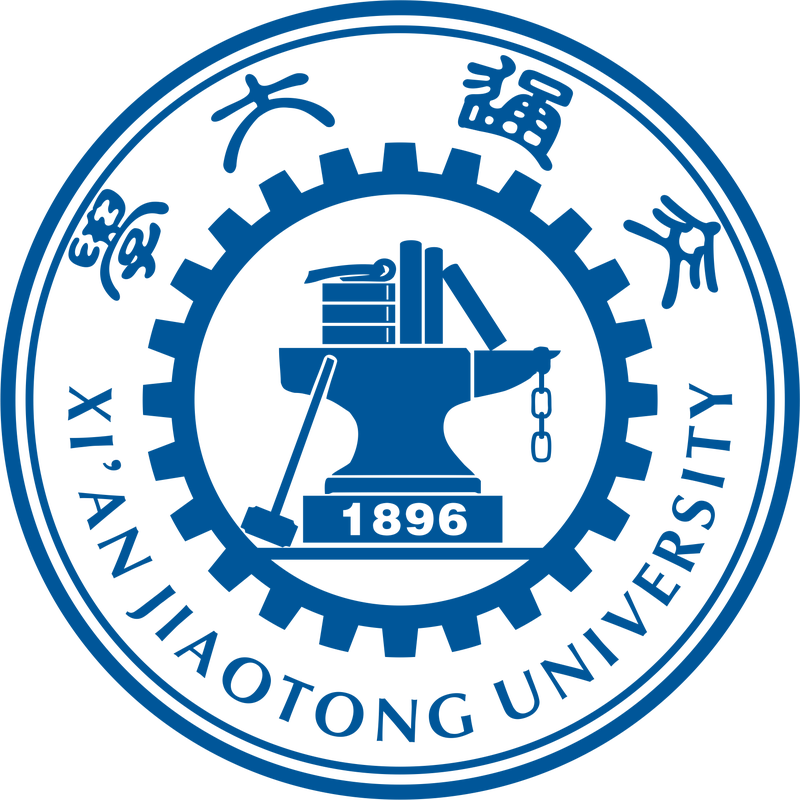}\hspace{4mm}%
    \includegraphics[height=12mm]{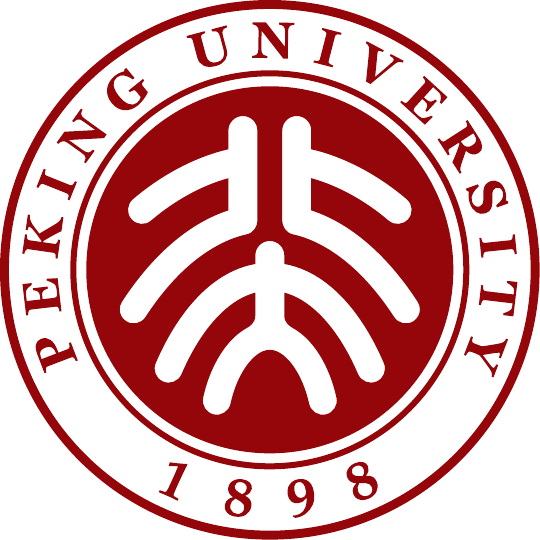}\hspace{4mm}%
    \includegraphics[height=12mm]{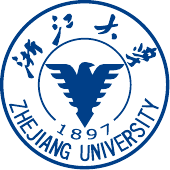}\hspace{4mm}%
    \includegraphics[height=12mm]{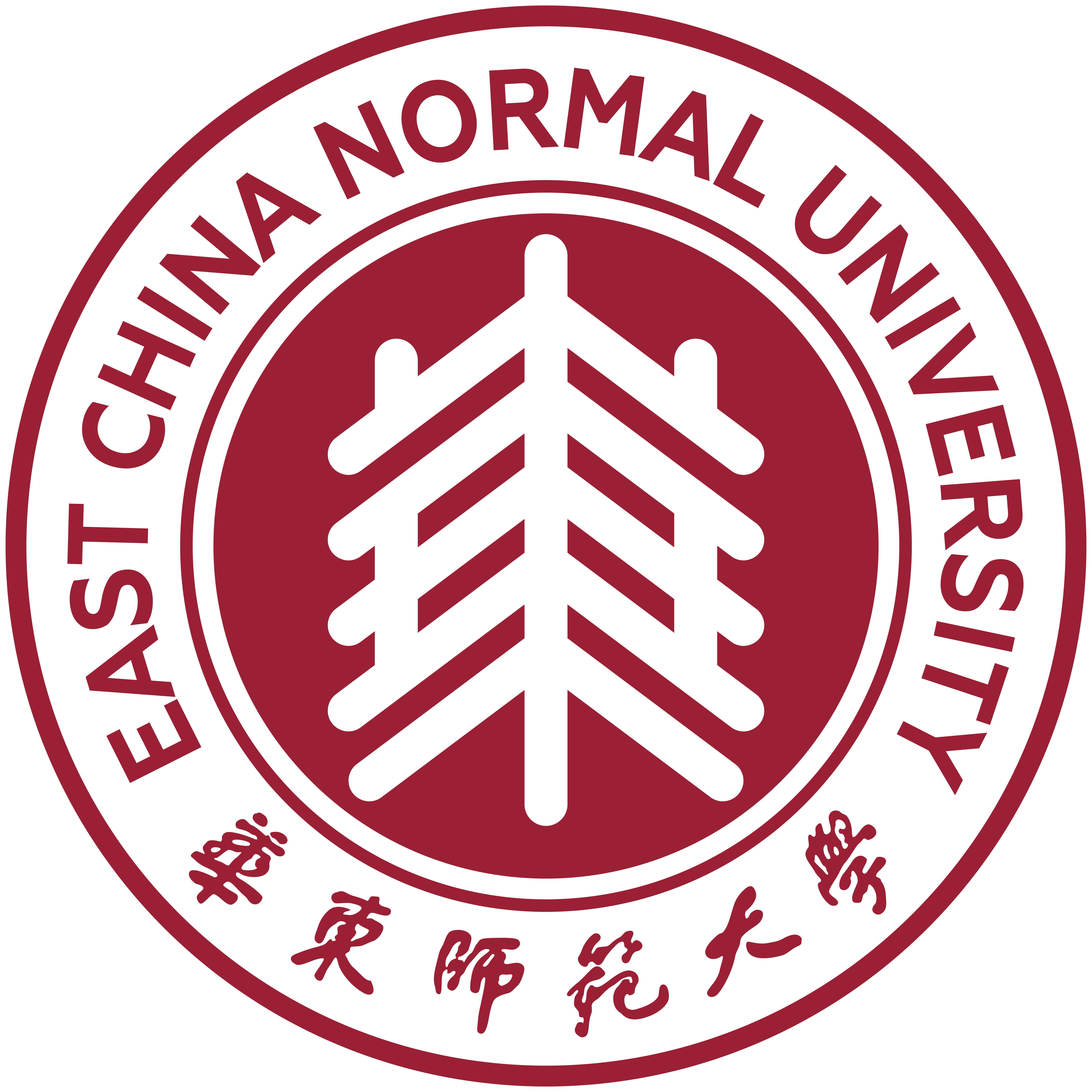}\hspace{4mm}%
  }%
}

\title{Do VLMs Need to Keep Looking at Long Videos\\While They Reason?\\Write It Down, Then Look Away}
\hypersetup{pdftitle={Do VLMs Need to Keep Looking at Long Videos While They Reason? Write It Down, Then Look Away}}

\author[1,*]{Pengyu Li}
\author[1,*]{Zhitao Gao}
\author[1,\dagger]{Lingling Zhang}
\author[1]{Heng Wang}
\author[2]{Chengqian Ma}
\author[3]{WenHao Wang}
\author[4]{Wenxuan Huang}
\author[5]{Chen Zheng}
\author[1]{Jun Liu}

\affiliation[1]{Xi'an Jiaotong University}
\affiliation[2]{Peking University}
\affiliation[3]{Zhejiang University}
\affiliation[4]{East China Normal University}
\affiliation[5]{Open University of China}

\contribution[*]{Equal contribution}
\contribution[\dagger]{Corresponding author}
\hypersetup{pdfauthor={\authorlist}}

\abstract{Vision-language models (VLMs) reason about long videos over hundreds to thousands of tokens.
Because every decoding step re-reads the video's key--value (KV) cache, memory-bandwidth traffic and decode latency scale with both video and reasoning length.
Must the model keep looking while it reasons?
In \method, the model first writes timestamped notes on what the question needs and puts the video away, so the inference engine hides its KV cache for the rest of the response.
Without training, \method nearly matches full-video thinking accuracy on Qwen3.6-27B at \N{e1.vh.notes.t}--\N{e1.mmmu.notes.t}\x{} its modeled decode cost.
Controlled probes across hybrid and pure-attention VLMs show why: visual dependence is concentrated early, and post-note reasoning relies on the recorded notes.
Replacing them with another video's notes sharply hurts accuracy, whereas keeping the video visible after note-taking adds no significant benefit.
To shorten this initial look, we fine-tune on the model's own correct \method responses and apply reinforcement learning that rewards correct answers while penalizing KV reads.
The trained model puts the video away after a median of \N{e45.vh.rlnotes.hidden}--\N{e45.mmmu.rlnotes.hidden} tokens.
Across three held-out benchmarks, it writes more tokens than an identically trained full-video model yet reads only \N{test.mmmu.reads.rlnotes.rlvan.pct}--\N{test.vh.reads.rlnotes.rlvan.pct}\% as many KV positions at matched accuracy.
}
\checkdata[Code]{\url{https://github.com/TiezMind/LookAway}}
\checkdata[Contact email]{\email{lipengyu.tiez@stu.xjtu.edu.cn}}

\begin{document}
\maketitle

\begin{figure}[t]
\centering
\includegraphics[width=\textwidth]{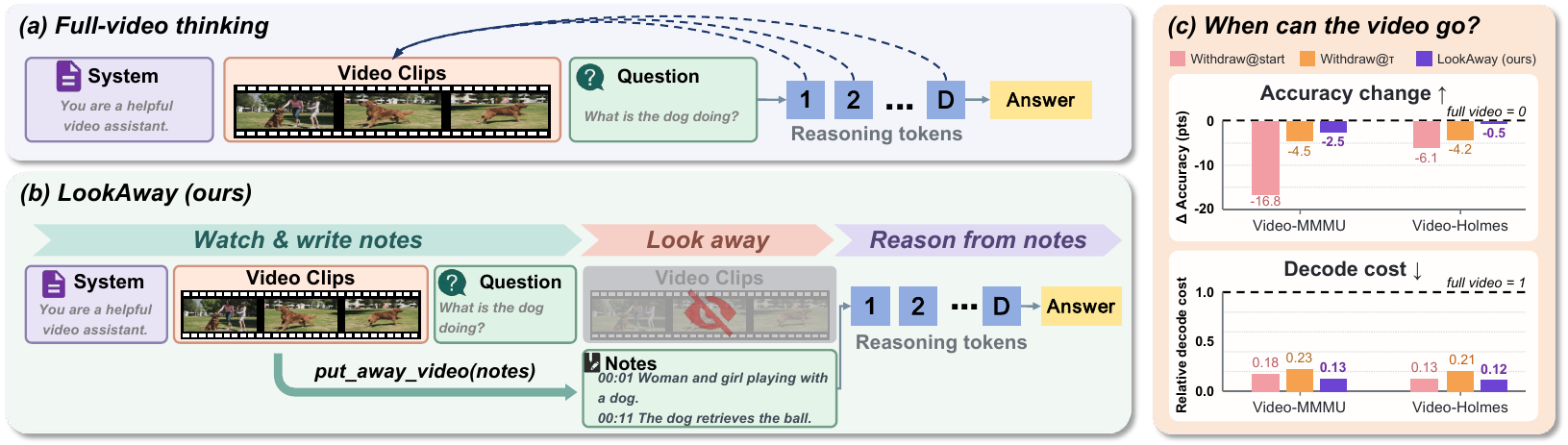}
\caption{\textbf{Do VLMs need to keep looking at long videos while they reason?} (a)~Every reasoning
token attends to the whole video. (b)~\method hides the video once the notes are written. (c)~Accuracy and modeled
decode cost (\cref{sec:cost}) of zero-shot Qwen3.6-27B on development splits, relative to full-video
thinking~(a). Withdraw@start hides the video before any tokens are generated. Withdraw@$\tau$ hides it at the same
step as \method after free reasoning instead of notes.}
\label{fig:teaser}
\end{figure}

\section{Introduction}
\label{sec:intro}

Vision-language models (VLMs) answer video questions through step-by-step reasoning that spans hundreds to thousands of tokens \cite{o1card,deepseekr1,qwen3vl}.
On long videos, every generated token attends to the key--value (KV) cache of the entire video.
Across our benchmarks, this cache holds a median of \N{prompt.vh.median} to \N{prompt.lvb.median} prompt tokens.
Decoding cost therefore grows with the product of video length and reasoning length.

Must a model keep reading the video throughout its reasoning?
Prior work points in opposite directions.
One line of work finds that long chains of thought lose track of visual evidence.
These methods re-inject or revisit visual inputs during reasoning \cite{tvc,reflectionv,lookback}.
Another line drops visual tokens in deeper layers, after a fixed decoding step, or during thinking, yet reports little accuracy loss \cite{vtw,fewereyes,lat}.
These works vary when visual tokens disappear, but they do not track what evidence the model has already written down.
Once a model records the relevant evidence in text, can it reason without the video?

People who reason about a lecture or a film rarely replay it while they think.
Instead, they write notes and reason from that external record \cite{risko2016}.
We give a VLM the same option through \method (\cref{fig:teaser}).
The response begins with a \texttt{put\_away\_video(notes)} tool call that records timestamped evidence for the question.
Once the call closes, the inference engine hides the video's KV blocks from attention.
The model then completes its reasoning using only its notes and the question.
Because withdrawal occurs at an explicit step, \method also serves as a controlled probe.
We independently vary when the video disappears, whether clips remain visible, and what text precedes withdrawal.
Zero-shot experiments on Qwen3.6-27B \cite{qwen36} yield three findings.
\begin{enumerate}
\item \textbf{Visual access is front-loaded.}
Hiding the video from the start causes a sharp accuracy drop, whereas keeping it visible only during initial decoding steps preserves full-video accuracy (\cref{sec:timing}).
\item \textbf{Notes match full-video reasoning at low cost.}
Writing brief notes lets \method hide the video after a median of \N{e1.mmmu.notes.hidden}--\N{e1.vh.notes.hidden} tokens, nearly matching full-video thinking accuracy at \N{e1.vh.notes.t}--\N{e1.mmmu.notes.t}\x{} its modeled decode cost while outperforming free reasoning withdrawn at the same step (\cref{fig:teaser}c).
\item \textbf{Post-note reasoning relies on the notes; keeping the video visible adds little.}
Replacing the notes with those from another video sharply reduces accuracy (\cref{fig:notes}), whereas keeping the video or selected clips visible after note-taking adds no significant accuracy gain (\cref{sec:pixels}).
\end{enumerate}

While writing these zero-shot notes, each step still reads the video.
To shorten this look, we train note-taking and withdrawal timing jointly.
Supervised fine-tuning on the model's own correct \method responses warms up the note format.
Reinforcement learning then rewards correct answers while charging each token for the KV reads of later steps.
Closing the tool call earlier removes the video from all later steps and lowers this charge.

\noindent We make three contributions.
\begin{itemize}
\item We introduce \method, a protocol for recording question-specific notes before hiding the video's KV cache.
An inference-engine plugin supports both thinking and non-thinking modes (\cref{sec:method}).
\item We show that dependence on the video is concentrated early in both a hybrid and a pure-attention VLM.
Subsequent answers rely primarily on the notes (\cref{sec:timing,sec:pixels}).
\item We train note-taking and withdrawal timing jointly with self-distillation SFT and read-cost RL (\cref{sec:train}), cutting KV reads through cheaper steps rather than shorter reasoning (\cref{sec:test}).
\end{itemize}

\section{Related Work}
\label{sec:related}

\paragraph{Visual evidence during reasoning.}
Multimodal reasoning models attend less to visual evidence as their chains of thought grow.
Several works therefore re-inject visual tokens or prompt the model to revisit the image \cite{tvc,reflectionv,lookback}.
Other methods prune visual tokens across layers or decoding steps \cite{fastv,vtw,visionpulse,dycoke,sparsevila}, or withdraw them after a fixed number of generated tokens \cite{fewereyes}.
Related approaches exclude visual tokens from thinking steps \cite{lat}, reason through a latent state \cite{cvrr}, or drop all visual KV entries during decoding \cite{twinprune}.
These works establish what visual tokens can be removed or restored.
On long videos, we separate withdrawal timing from the text written before withdrawal.
Once the model writes down the needed evidence, re-reading the video adds little.

\paragraph{Text as working memory.}
Many video systems answer questions from clip captions \cite{llovi} or agent transcripts that replace the frames \cite{omniagent,sprout,mmcontextfold}.
Recent methods also reward models when their generated descriptions suffice to answer the question \cite{visionsr1,visionaryr1}.
However, written text replaces visual KV entries only for the facts it states \cite{seensaid,li2026visual,gao2026aero,li2026agtao, li2026lookcloserlookback}.
Language models similarly compress or summarize their context during reasoning \cite{lightthinker,inftythink,pencil,mem1,memagent,statelm}.
In \method, one model writes question-specific notes and reasons from them within a single response and prefill.
Controlled interventions confirm that the answer follows these notes.

\paragraph{Efficient reasoning and KV reads.}
Length penalties and token budgets shorten chains of thought \cite{l1,thinkprune,laser,overthinking}.
KV eviction and sparse attention instead reduce what each decoding step reads \cite{h2o,snapkv,quest,rkv}.
Declarative Attention lets a language model select the document chunks it attends to \cite{da}.
While KV eviction retains entries with high attention scores, \method lets the model record the needed evidence as text before hiding the video.
When the video dominates the KV cache, hiding it lowers the cost of each decoding step rather than only the number of steps.
We train this behavior by penalizing KV reads instead of response length.

\section{Preliminaries}
\label{sec:prelim}

\subsection{Long-Video Reasoning}
\label{sec:setting}

A request to a VLM consists of a prompt prefix $H$, a video $V$, and a question $Q$.
The prefix contains the system message, any tool schema, and a line introducing the clips.
The question includes the answer options and task instructions.
The video arrives as $M$ time-ordered clips $V=(V_1,\dots,V_M)$, each labeled with its time range.
Prefill encodes the prompt $x=(H,V,Q)$ once, and the model then generates a response $y=(y_1,\dots,y_D)$.
The response contains reasoning in a thinking block or in the reply itself, followed by an answer.
Let $\mathcal{H}$, $\mathcal{V}$, and $\mathcal{Q}$ denote the prompt positions of $H$, $V$, and $Q$, and let $\mathcal{G}_s$ denote the positions of $y_{<s}$.
At decoding step $s$, attention reads the KV entries of a visible set $\mathcal{K}_s$.
Ordinary decoding keeps all positions visible, so $\mathcal{K}_s=\mathcal{H}\cup\mathcal{V}\cup\mathcal{Q}\cup\mathcal{G}_s$.
A rule that updates $\mathcal{K}_s$ from the generated text defines a \emph{visual access protocol}.
This protocol leaves prefill unchanged while controlling what each step reads and how long the response runs.

We study Qwen3.6-27B \cite{qwen36}, a natively multimodal model with two reasoning modes.
Thinking mode places reasoning in a thinking block before the reply, whereas non-thinking mode reasons directly in the reply.
All experiments use thinking mode unless noted.
In this hybrid architecture, 16 of the 64 layers use full attention and maintain the KV cache.
The remaining 48 Gated DeltaNet layers \cite{gateddeltanet} maintain a fixed-size recurrent state, so only the full-attention layers read $\mathcal{K}_s$.
We sample up to 2 frames per second and cap each video at 768 frames of 230{,}400 pixels, grouped into 16-frame clips.
Video tokens dominate the prompt, with a median of \N{prompt.vh.median} tokens on Video-Holmes and \N{prompt.mmmu.median}--\N{prompt.lvb.median} on VideoMMMU and LVBench.

\subsection{Decoding Cost}
\label{sec:cost}

Because prefill is identical across protocols, we compare protocols by their decoding cost.
Let $A=\sum_{s=1}^{D}|\mathcal{K}_s|$ be the KV positions read by the full-attention layers during decoding.
Under the roofline model of \cite{roofline,da} for batched decoding, weight reads are shared across the batch.
The modeled decode time of a request is
\begin{equation}
T \;=\; D\left(\frac{2N}{F\mu_F} + \frac{s_{\mathrm{loc}}}{W\mu_W}\right) \;+\; \frac{A\,b_{\mathrm{kv}}}{W\mu_W}.
\label{eq:cost}
\end{equation}
Here $N=27$B is the parameter count, $b_{\mathrm{kv}}=64$\,KiB is the KV size per position, and $s_{\mathrm{loc}}$ is the recurrent state read per step.
Constants $F$ and $W$ are the peak compute and bandwidth of a B200 GPU at utilizations $\mu_F=0.40$ and $\mu_W=0.70$ (\cref{app:cost}).
Decode time thus depends on the step count $D$ and the visible set size $|\mathcal{K}_s|$.
Shorter reasoning reduces $D$, whereas hiding the video reduces $|\mathcal{K}_s|$.
With a 45K-token video visible, KV reads approaches 90\% of a step.
We report cost and KV reads as ratios of summed $T$ and summed $A$ to a reference on the same questions.

\section{Method}
\label{sec:method}

\begin{figure}[t]
\centering
\includegraphics[width=\textwidth]{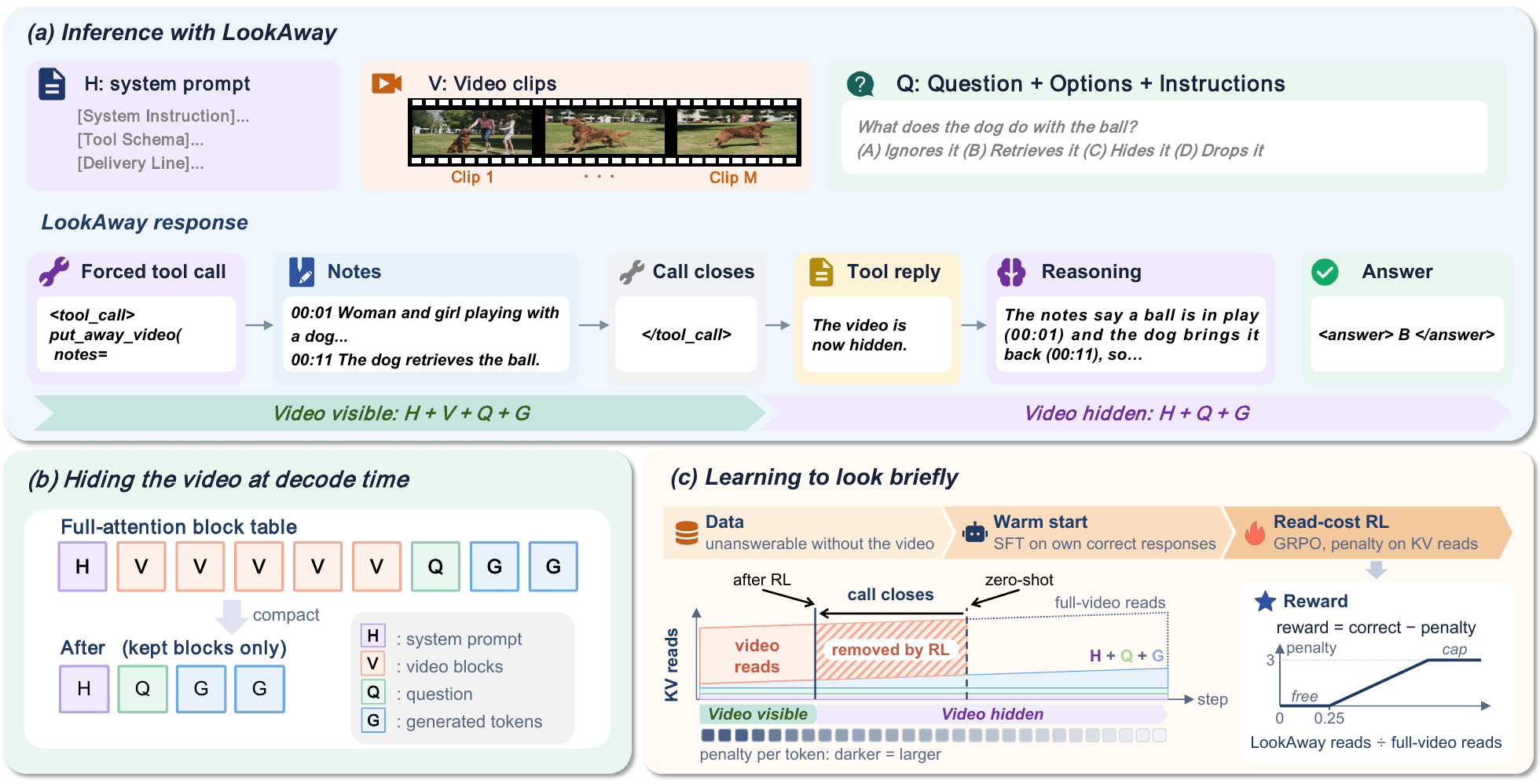}
\caption{\textbf{\method.} In (c), the chart sketches the KV reads of each decoding step. The penalty
  grows with the response's reads relative to full-video thinking (the solid area divided by the area under
  the dotted line). Each token pays only for the reads of subsequent steps up to the response penalty, so the notes
  pay most. Both training stages hide the video as in (b).}
\label{fig:method}
\end{figure}

\method lets a VLM read a long video only while writing down the evidence needed for the question (\cref{fig:method}).
The protocol defines how the model records notes and puts the video away (\cref{sec:protocol}).
The inference engine then hides the video KV blocks during decoding (\cref{sec:engine}).
Finally, read-cost training teaches the model to gather evidence and stop looking earlier (\cref{sec:train}).

\subsection{Write It Down, Then Look Away}
\label{sec:protocol}

The prompt defines a single tool, \texttt{put\_away\_video}, that takes notes as a string argument.
The instructions ask the model to record question-relevant evidence from the video with one timestamped entry per line.
The model then answers from its notes and the question alone.
Compared with the full-video prompt, only the tool definition and this instruction differ.

The response begins with forced tool-call tokens, preceded in thinking mode by an empty thinking block.
While the video remains visible, the model writes its notes and chooses when to close the call.
A fixed tool response then confirms that the video is hidden.
Within the same response, the model continues reasoning (opening a new thinking block in thinking mode) and produces its final answer.
Let $\tau$ denote the step at which the call closes.
The set of visible KV positions at decoding step $s$ is
\begin{equation}
\mathcal{K}_s=\begin{cases}
\mathcal{H}\cup\mathcal{V}\cup\mathcal{Q}\cup\mathcal{G}_s, & s\le\tau,\\
\mathcal{H}\cup\mathcal{R}\cup\mathcal{Q}\cup\mathcal{G}_s, & s>\tau,
\end{cases}
\label{eq:access}
\end{equation}
where $\mathcal{V}$ contains all video token positions and $\mathcal{R}\subseteq\mathcal{V}$ contains any clips retained after withdrawal.
\method sets $\mathcal{R}=\varnothing$, whereas experimental controls use nonempty subsets (\cref{sec:controls}).
The notes belong to $\mathcal{G}_s$ rather than $\mathcal{R}$.
After withdrawal, \method attends only to the prompt prefix, the question, and the generated history.
Retained spans are rounded to KV-block boundaries (\cref{sec:engine}), and visual access is never restored.

Two properties distinguish \method from prior visual-token withdrawal methods.
First, the model writes question-specific notes and chooses the withdrawal step $\tau$ through the tool call.
Prior methods instead withdraw visual tokens at a fixed step \cite{fewereyes} or by attention scores \cite{snapkv,rkv}.
Second, visual access stays disabled for the rest of the response, unlike methods that restore it in later steps \cite{lat}.
The notes serve as an explicit record for subsequent reasoning.
The controls in \cref{sec:controls} separate the effects of withdrawal timing and note content.

\subsection{Hiding the Video at Decode Time}
\label{sec:engine}

An engine plugin for vLLM \cite{vllm} implements \cref{eq:access} via a per-request state machine over generated tokens.
Once the call closes, the plugin rebuilds the attention metadata for the full-attention layers.
It compacts the block table to the kept 32-token KV blocks (rounding spans outward) and shortens sequence lengths accordingly.
The attention kernel therefore reads only the kept blocks.
Because the KV cache itself is untouched, prefix caching across requests is unaffected and prefill is unchanged.
Single-prefill execution also leaves the Gated DeltaNet state and the video-conditioned question and note KV entries intact.
\Cref{sec:pixels} tests whether these residual channels carry the answer.
The plugin also counts the KV positions read per step to supply the cost signal for training (\cref{app:engine}).

\subsection{Learning to Look Briefly}
\label{sec:train}

Zero-shot, the model writes detailed notes and closes the call after a median of \N{e1.mmmu.notes.hidden}--\N{e1.vh.notes.hidden} tokens.
We train it to look briefly in two stages.
To isolate the protocol, we train a \emph{full-video RL} counterpart on the same questions with the same warm-start procedure, recipe, and reward.

\paragraph{Warm start.}
The base model answers each training question twice under each protocol.
We keep one correct response per question on the 715 questions solved under both protocols, and fine-tune for one epoch with the visibility of \cref{eq:access} applied in the forward pass.

\paragraph{Read-cost RL.}
We then run GRPO \cite{deepseekmath} with the visibility of \cref{eq:access} applied in rollouts and policy updates.
Let $a_{i,s}$ be the KV positions read at step $s$ of response $i$, and $A_i$ their sum plus a per-question prompt term \cite{da}.
A response with correctness $r_i\in\{0,1\}$ earns
\begin{equation}
R_i = r_i - E_i,\quad E_i=\min\!\big(\kappa,\ \max(0,\ A_i/a_{\mathrm{ref}} - f)\big),
\label{eq:reward}
\end{equation}
where $a_{\mathrm{ref}}$ is the mean read count of the base model's full-video responses on the same question.
Here $f=0.25$ is a free budget and $\kappa=3$ caps the charge $E_i$.
GRPO gives every token of a response the same advantage, which would penalize post-withdrawal reasoning for the reads of the notes.
We instead charge each token its \emph{cost-to-go}, the reads of all later steps capped at $E_i$.
The per-token charge and advantage are
\begin{equation}
\begin{aligned}
c_{i,t}&=\min\!\Big({\textstyle\sum_{s>t}}\,a_{i,s}/a_{\mathrm{ref}},\ E_i\Big),\\
\hat A_{i,t}&=(r_i-\bar r)-(c_{i,t}-\bar c_t),
\end{aligned}
\label{eq:adv}
\end{equation}
where $\bar r$ and $\bar c_t$ are group means (\cref{app:training}).
Closing the call one step earlier removes the video from every later step and lowers the cost-to-go before $\tau$.
The saving is therefore credited to the notes, where the model decides to stop looking.
After the call every step is cheap, so reasoning is judged mainly by correctness.

\section{Experimental Setup}
\label{sec:exp}

\subsection{Data and Evaluation}
\label{sec:setup}

\paragraph{Benchmarks and splits.}
Our multiple-choice benchmarks cover short-film reasoning (Video-Holmes \cite{videoholmes}), lectures (VideoMMMU \cite{videommmu}), and hour-long videos (LVBench \cite{lvbench}).
For VideoMMMU, we evaluate the Perception and Comprehension tracks.
The zero-shot analyses in \cref{sec:timing,sec:pixels} use two development splits across two seeds.
VideoMMMU-dev contains 200 questions on 100 videos, and Video-Holmes-dev contains 661 questions on 100 films from the training release.
Two additional checks use LVBench-dev with 312 questions on 20 videos and one seed.
The held-out splits in \cref{sec:test} share no videos with these development splits and serve only for final evaluation.

\paragraph{Training data.}
Training questions come from the 2--3-minute academic subset of LLaVA-Video-178K \cite{llavavideo} and the remaining Video-Holmes training films.
To require visual evidence, we exclude questions that the base model answers without the video in at least two of four attempts.
For RL, we also exclude questions solved by all four base-model rollouts, leaving 542 questions for 150 training steps in verl~\cite{hybridflow}.
Both the full-video and \method models use the exact same questions and training recipe (\cref{app:training}).

\paragraph{Metrics and uncertainty.}
Accuracy is averaged over seeds and questions.
Differences ($\Delta$) report the named condition minus its reference in percentage points, and cost denotes the modeled decode-time ratio from \cref{sec:cost}.
A difference is statistically significant when its 95\% paired-bootstrap interval over videos excludes zero (\cref{app:intervals,app:full}).

\subsection{Experimental Controls}
\label{sec:controls}

Controlled variants of \cref{eq:access} isolate when the model needs the video and what information supports subsequent reasoning.
On the same questions, we independently vary the withdrawal step $\tau$, the retained clips $\mathcal{R}$, and the text written before withdrawal.
Full-video thinking keeps the video visible throughout ($\tau=\infty$).
\begin{itemize}
\item \textbf{When the video goes.}
\emph{Withdraw@$\tau$} uses the full-video prompt and hides the video without warning at $\tau$, the seed-averaged step where \method closes its call.
The model thus writes free reasoning rather than notes before withdrawal.
\emph{Withdraw@512} hides the video after 512 tokens, whereas \emph{Withdraw@start} forces an empty call and hides the video before any tokens are generated.
\emph{Notes + video} keeps the \method prompt, call, and tool reply unchanged, but leaves the video readable after $\tau$.
\item \textbf{What stays visible.}
\emph{Notes + clips} retains up to four model-selected clips in $\mathcal{R}$ after note-taking.
Alternative controls select the same number of clips using SigLIP \cite{siglip} query similarity or random sampling.
\emph{Clips only} forces empty notes while retaining the model-selected clips.
\item \textbf{What the notes say.}
\emph{Forced notes} pre-fills the tool call with specified notes and then hides the video.
We test the model's own notes, line-shuffled notes, or notes from another video in the same benchmark.
\end{itemize}
Together, these controls instantiate key baseline families within the same base model.
Withdraw@$\tau$ mimics silent KV eviction, Withdraw@512 matches fixed-step withdrawal \cite{fewereyes}, and SigLIP clip retention represents query-aware frame selection.
During training, full-video RL (\cref{sec:train}) reads the video at every step, so its read penalty acts as a response-length penalty \cite{l1,thinkprune}.
\Cref{app:engine} details all prompts, controls, and non-thinking mode configurations.

\section{Results and Analysis}
\label{sec:results}

\subsection{Timing of Visual Access}
\label{sec:timing}

\begin{table}[t]
\centering
\small
\setlength{\tabcolsep}{4pt}
\resizebox{\textwidth}{!}{\begin{tabular}{@{}ll>{\footnotesize}l>{\footnotesize}l rrrr rrrr@{}}
\toprule
 & & \multicolumn{2}{c}{\footnotesize Video} & \multicolumn{4}{c}{VideoMMMU-dev} & \multicolumn{4}{c}{Video-Holmes-dev} \\
\cmidrule(lr){3-4}\cmidrule(lr){5-8}\cmidrule(l){9-12}
Condition & & withdrawn & written by then & Acc & $\Delta$ & Tokens & Cost & Acc & $\Delta$ & Tokens & Cost \\
\midrule
\multicolumn{12}{@{}l}{\textbf{Thinking mode}} \\
\addlinespace[1pt]
\multicolumn{12}{@{}l}{\emph{Reference}} \\
\quad Full video &  & \color{tabgray}never & \color{tabgray}-- & 85.8 & -- & 3,425 & 1 & 71.3 & -- & 3,162 & 1 \\
\midrule
\multicolumn{12}{@{}l}{\emph{Training-free baselines}} \\
\quad Budget forcing~\cite{s1} & $N{=}1024$ & \color{tabgray}never & \color{tabgray}-- & \textbf{86.2} & +0.5 & 1,779 & 0.52 & 70.5 & $-$0.8 & 1,423 & 0.43 \\
 & $N{=}512$ & \color{tabgray}never & \color{tabgray}-- & 84.5 & $-$1.3 & 1,544 & 0.44 & 70.7 & $-$0.7 & 1,091 & 0.33 \\
 & $N{=}256$ & \color{tabgray}never & \color{tabgray}-- & 83.5 & $-$2.3 & 1,349 & 0.39 & 71.1 & $-$0.2 & 946 & 0.29 \\
\quad Fixed-step withdrawal~\cite{fewereyes} & $K{=}512$ & \color{tabgray}after $K$ & \color{tabgray}reasoning & \underline{85.8} & \underline{0.0} & 3,343 & 0.26 & \textbf{71.7} & +0.4 & 2,877 & 0.29 \\
 & $K{=}256$ & \color{tabgray}after $K$ & \color{tabgray}reasoning & 80.8 & $-$5.0 & 3,798 & 0.22 & 67.9 & $-$3.4 & 2,887 & 0.23 \\
 & $K{=}128$ & \color{tabgray}after $K$ & \color{tabgray}reasoning & 75.8 & $-$10.0 & 4,517 & 0.23 & 68.1 & $-$3.3 & 2,728 & 0.18 \\
 & $K{=}64$ & \color{tabgray}after $K$ & \color{tabgray}reasoning & 71.5 & $-$14.3 & 5,344 & 0.26 & 66.7 & $-$4.6 & 2,572 & 0.16 \\
\midrule
\multicolumn{12}{@{}l}{\emph{Access controls}} \\
\quad Notes + video &  & \color{tabgray}never & \color{tabgray}notes & 84.5 & $-$1.3 & 2,132 & 0.62 & 69.8 & $-$1.5 & 1,596 & 0.50 \\
\quad Withdraw@$\tau$ &  & \color{tabgray}at $\tau$ & \color{tabgray}reasoning & 81.2 & $-$4.5 & 4,203 & 0.23 & 67.1 & $-$4.2 & 2,709 & 0.21 \\
\quad Clips only &  & \color{tabgray}at start$^\ast$ & \color{tabgray}clip names & 78.5 & $-$7.3 & 3,364 & 0.21 & 66.9 & $-$4.4 & 2,407 & 0.24 \\
\quad Withdraw@start &  & \color{tabgray}at start & \color{tabgray}nothing & 69.0 & $-$16.8 & 4,073 & 0.18 & 65.3 & $-$6.1 & 1,858 & 0.13 \\
\midrule
\multicolumn{12}{@{}l}{\emph{Ours}} \\
\rowcolor{oursrow}\quad \method &  & \color{tabgray}at $\tau$ & \color{tabgray}notes & 83.2 & $-$2.5 & 2,025 & \textbf{0.13} & \underline{70.9} & \underline{$-$0.5} & 1,199 & \textbf{0.12} \\
\midrule
\multicolumn{12}{@{}l}{\textbf{Non-thinking mode}} \\
\quad Full video &  & \color{tabgray}never & \color{tabgray}-- & 84.2 & -- & 774 & 1 & 71.0 & -- & 323 & 1 \\
\rowcolor{oursrow}\quad \method &  & \color{tabgray}at $\tau$ & \color{tabgray}notes & 85.2 & +1.0 & 696 & 0.30 & 69.6 & $-$1.4 & 436 & 0.67 \\
\bottomrule
\end{tabular}
}
\caption{\textbf{When must the model look?} Zero-shot Qwen3.6-27B, two seeds.
$N$ is the thinking budget and $K$ is the fixed withdrawal step.
$\tau$ marks where \method closes its call on the same question.
$^\ast$ indicates retained model-selected clips.
$\Delta$ (points) and Cost (modeled decode time) are relative to full video in the same mode.
Tokens reports mean response length. In thinking mode, bold marks best accuracy and lowest cost.
Underlining marks the cheapest condition within one point of full-video accuracy.}
\label{tab:access}
\end{table}

\finding{Visual access is front-loaded.}{The video matters at the start of the response and little after it.}

Withdraw@start loses \N{e1.mmmu.blindloss} points on VideoMMMU and \N{e1.vh.blindloss} on Video-Holmes (\cref{tab:access}).
By contrast, Withdraw@512 matches or slightly exceeds full-video thinking in point estimate while reducing decode cost.
Initial visual access thus matters more than keeping the video visible throughout.

The same pattern holds after explicit note-taking.
Notes + video keeps the video readable after the call, yet adds no significant accuracy over \method while increasing decode cost (\cref{tab:access,app:intervals}).
Front-loading also appears in Qwen3-VL-32B-Thinking~\cite{qwen3vl}, where every layer uses full attention.
Hiding the video at the start hurts accuracy, whereas hiding it at the note-taking step $\tau$ after free reasoning changes accuracy little in point estimate (\cref{app:qvl}).

\finding{Notes match full-video reasoning at low cost.}{\method nearly matches full-video accuracy at
\N{e1.vh.notes.t}--\N{e1.mmmu.notes.t}\x{} its cost, outperforming matched-step free reasoning.}

\method withdraws the video after a median of \N{e1.mmmu.notes.hidden} and \N{e1.vh.notes.hidden} response tokens, staying within \N{x.vh.notes-vanilla.abs}--\N{x.mmmu.notes-vanilla.abs} points of full-video thinking accuracy at \N{e1.vh.notes.t}--\N{e1.mmmu.notes.t}\x{} its modeled decode cost (\cref{fig:teaser}c, \cref{tab:access}).
Among the thinking variants, \method is the cheapest: it costs less than every fixed-step withdrawal baseline and is more accurate than those withdrawing before step 512.

To separate note content from withdrawal timing, we compare against Withdraw@$\tau$, which hides the video at the same step after free reasoning under the full-video prompt.
At this matched step, \method gains \N{e1.mmmu.primary.abs} points on VideoMMMU and a significant \N{e1.vh.primary.abs} points on Video-Holmes (\cref{app:intervals}).
Without notes, the model writes roughly twice as many tokens, searching for details it can no longer see (``let me re-watch'', \cref{fig:qual}a).
Notes help most on films, whereas early visual access alone suffices on lecture videos (\cref{tab:access}).

Closing the thinking block is distinct from withdrawing the video.
With thinking budgets of 256 to 1024 tokens~\cite{s1}, the model keeps reasoning in its reply while every step reads the video.
Even the shortest tested budget costs \N{bd.vh.bf256.vsnotes}--\N{bd.mmmu.bf256.vsnotes}\x{} as much as \method at similar accuracy (\cref{tab:access}).

\paragraph{Non-thinking mode.}
The same protocol applies when the model reasons in its reply.
On VideoMMMU and Video-Holmes, \method costs \N{eff.mmmu.ntnotes.t}--\N{eff.vh.ntnotes.t}\x{} as much as full-video non-thinking mode (\cref{tab:access}, bottom).
Accuracy changes by \N{eff.mmmu.ntnotes.d} and \N{eff.vh.ntnotes.d} points, and savings grow on longer replies (\cref{app:engine}).

\subsection{Notes versus Pixels}
\label{sec:pixels}

\begin{figure}[t]
\begin{sidecaption}{0.58}{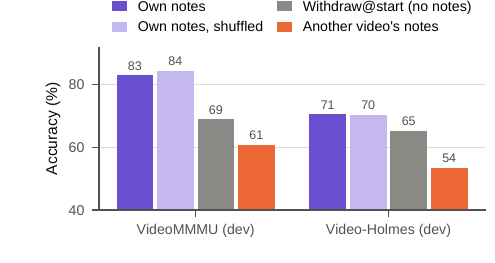}
\caption{\textbf{The answer follows what the notes say.} We pre-fill the tool call with the given notes and then
hide the video. Results are zero-shot across two seeds.}
\label{fig:notes}
\end{sidecaption}
\end{figure}

If subsequent reasoning relies on the notes, changing their content should change the answer.
We test this by pre-filling the call with given notes before hiding the video (\cref{fig:notes}).
Because single-prefill execution also leaves video information in the recurrent state and in the question and note KV entries, we isolate the pure text contribution using a fresh prompt that never held the video.

\finding{Post-note reasoning relies on the notes; keeping the video visible adds little.}{Wrong notes hurt more than
none, while keeping video or clips visible adds no significant gain.}

\paragraph{What the notes say.}
Replaying the model's own notes closely reproduces free-generation accuracy.
Replacing them with another video's notes reduces accuracy by \N{e45.mmmu.own-swap.abs} points on VideoMMMU and \N{e45.vh.own-swap.abs} on Video-Holmes.
These wrong notes perform worse even than Withdraw@start, which writes nothing (\cref{fig:notes}).
By contrast, shuffling the model's own notes has no significant effect (\cref{app:intervals}).
Note content, rather than ordering, therefore drives the answer.

The note text carries most of this benefit on its own.
Given to a prompt that never held the video, the model's notes retain \N{e2.vh.textshare}--\N{e2.mmmu.textshare}\% of their accuracy gain over the question alone (\cref{app:e2}).
\method exceeds this text-only re-prompt because single-prefill execution also keeps video-conditioned representations.
These channels run primarily through the question tokens.
Blocking the question from attending to the clips during prefill reduces immediate-answer accuracy to the no-video level, even though the recurrent state still holds the video (\cref{app:mech}).

\paragraph{What the pixels add.}
Zero-shot, keeping up to four model-selected clips alongside the notes adds no significant accuracy over \method.
Selecting four clips via SigLIP similarity or at random makes no significant difference either (\cref{app:intervals}).
Keeping model-selected clips without notes (Clips only) falls well behind \method on both splits (\cref{tab:access}).

After training (\cref{sec:dynamics}), keeping clips likewise adds no significant accuracy while increasing decode cost.
The upper 95\% bounds on the accuracy gains are \N{e45.mmmu.RLpins-RLnotes.hi} and \N{e45.vh.RLpins-RLnotes.hi} points on VideoMMMU and Video-Holmes (\cref{app:intervals}).

\begin{figure}[t]
\begin{sidecaption}{0.58}{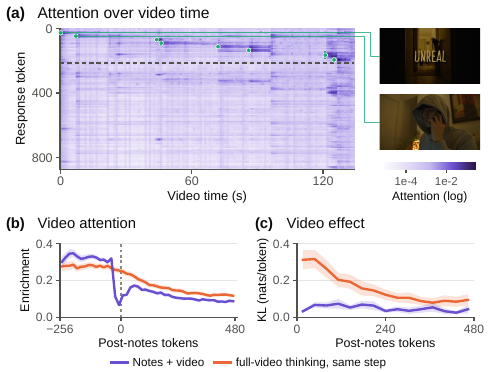}
\caption{\textbf{The model reads the video to write.}
(a) Token attention over video time for Notes + video, which retains visual access.
Green marks cited timestamps and the dashed line marks the end of the notes.
(b) Video attention relative to key share around the end of the notes (\N{mech.n} questions).
(c) KL divergence between next-token distributions with the video readable versus hidden.}
\label{fig:mech}
\end{sidecaption}
\end{figure}

\paragraph{Why the pixels add little.}
During note-taking, attention concentrates on the frames each note cites.
Hiding cited clips lowers note-word log-probability more than hiding as many random clips (\cref{fig:mech,app:mech}).

In Notes + video, where the video stays readable, attention to it drops by \N{mech.p5.drop}\% after the notes, compared with \N{mech.p5.fulldrop}\% at matched full-video steps.
Holding response text fixed, the KL divergence between readable-video and hidden-video next-token distributions after the notes is only \N{mech.p3.ratio}\x{} that of full-video thinking.

\subsection{Training Dynamics}
\label{sec:dynamics}

\begin{figure}[t]
\centering
\includegraphics[width=\textwidth]{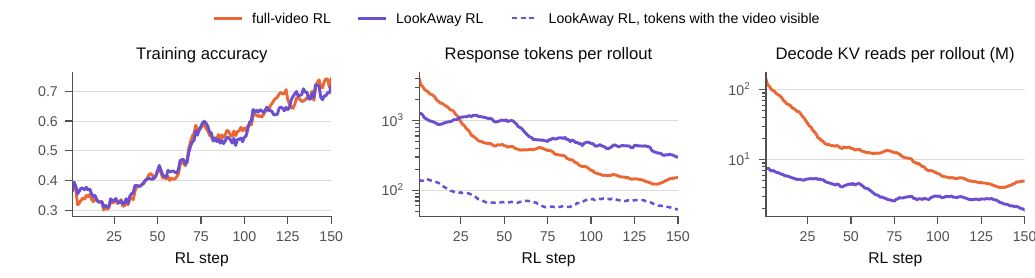}
\caption{\textbf{Both RL models learn at similar rates but save cost differently.} Curves are 10-step moving averages.}
\label{fig:training}
\end{figure}

\finding{Looking briefly can be learned.}{Read-cost RL shortens the look and the notes, and development accuracy
rises.}

Both models improve at similar rates during training but cut decode cost differently (\cref{fig:training}).
Full-video RL shortens responses while reading the video at each step; \method RL shortens them less and reads it only for notes.

On both development splits, read-cost RL cuts the median withdrawal step from \N{e1.mmmu.notes.hidden}--\N{e1.vh.notes.hidden} to \N{e45.vh.rlnotes.hidden}--\N{e45.mmmu.rlnotes.hidden} tokens, shortening the notes while raising accuracy (\cref{app:training}).

\subsection{Held-out Evaluation}
\label{sec:test}

We evaluate on the Video-Holmes test release (\N{test.vh.q} questions on \N{test.vh.videos} films), the remaining VideoMMMU videos (\N{test.mmmu.q} questions on \N{test.mmmu.videos} videos), and the LVBench test split (\N{test.lvb.q} questions on \N{test.lvb.videos} hour-long videos).
\Cref{tab:test} reports all models, and \cref{app:full} gives 95\% intervals.

\begin{table}[t]
\centering
\small
\begin{tabular}{@{}l rrr rrr rrr@{}}
\toprule
 & \multicolumn{3}{c}{Video-Holmes (in-domain)} & \multicolumn{3}{c}{VideoMMMU} & \multicolumn{3}{c}{LVBench} \\
\cmidrule(lr){2-4}\cmidrule(lr){5-7}\cmidrule(l){8-10}
Model & Acc & $\Delta$ & Cost & Acc & $\Delta$ & Cost & Acc & $\Delta$ & Cost \\
\midrule
Full-video thinking & 59.1 & -- & 1 & 88.2 & -- & 1 & \textbf{66.5} & -- & 1 \\
Full video, non-thinking & 58.6 & $-$0.5 & 0.102 & 86.8 & $-$1.5 & 0.200 & 66.3 & $-$0.3 & 0.060 \\
\method{}, zero-shot & 57.5 & $-$1.6 & 0.119 & 86.8 & $-$1.5 & 0.127 & 65.5 & $-$1.0 & 0.155 \\
\midrule
Full-video RL & 62.1 & +3.0 & 0.062 & \textbf{88.5} & \textbf{+0.2} & 0.119 & 63.6 & $-$2.9 & 0.052 \\
\rowcolor{oursrow}\method{} RL & \textbf{62.3} & \textbf{+3.2} & \textbf{0.036} & 87.0 & $-$1.2 & \textbf{0.047} & 63.7 & $-$2.8 & \textbf{0.023} \\
\bottomrule
\end{tabular}

\caption{\textbf{Held-out evaluation} of Qwen3.6-27B (two seeds). Acc is accuracy (\%).
$\Delta$ (points) and Cost (modeled decode time) are relative to full-video thinking.
Bold marks best accuracy and lowest cost per benchmark.
\Cref{tab:full} includes warm-start (SFT) checkpoints.}
\label{tab:test}
\end{table}

\finding{The saving comes from cheaper steps, not shorter reasoning.}{Held out, \method RL writes more tokens
than full-video RL, yet each step reads \N{test.lvb.stepreads.rlnotes.rlvan.x}--\N{test.mmmu.stepreads.rlnotes.rlvan.x}\x{}
fewer KV positions.}

\paragraph{Against the identically trained full-video model.}
Trained on the same questions, warm-start procedure, and RL recipe, \method RL writes \N{test.lvb.tokratio.rlnotes.rlvan}--\N{test.vh.tokratio.rlnotes.rlvan}\x{} as many tokens as full-video RL.
Most are decoded with the video hidden, so it reads only \N{test.mmmu.reads.rlnotes.rlvan.pct}--\N{test.vh.reads.rlnotes.rlvan.pct}\% as many KV positions.

Under the decode cost model, \method RL costs \N{test.mmmu.H3.t}--\N{test.vh.H3.t}\x{} as much as full-video RL.
Accuracy is statistically indistinguishable across all three benchmarks, with point estimates slightly higher on Video-Holmes and LVBench and within \N{test.mmmu.H3.d.abs} points on VideoMMMU (\cref{tab:test,app:full}).

\paragraph{Against untrained full-video thinking.}
\method RL decodes at \N{test.lvb.rlnotes.t}--\N{test.mmmu.rlnotes.t}\x{} the cost of untrained full-video thinking and gains \N{test.vh.rlnotes.d.abs} points on in-domain Video-Holmes.
It also costs under half as much as non-thinking full-video decoding on every benchmark (\cref{tab:test}).
These gains combine RL with the protocol, whereas the matched comparison above isolates it.

Out of domain, accuracy changes by \N{test.mmmu.rlnotes.d} points on VideoMMMU and \N{test.lvb.rlnotes.d} on LVBench.
Because both RL models are trained on 2--3-minute videos, full-video RL shows the same shift on hour-long LVBench (\N{test.lvb.rlvan.d}).
Without this training shift, zero-shot \method stays within \N{test.lvb.basenotes.d.abs} point of full-video thinking on LVBench (\cref{tab:test}), and its savings persist across smaller serving batches (\cref{app:cost}).

\begin{figure}[t]
\centering
\includegraphics[width=\textwidth]{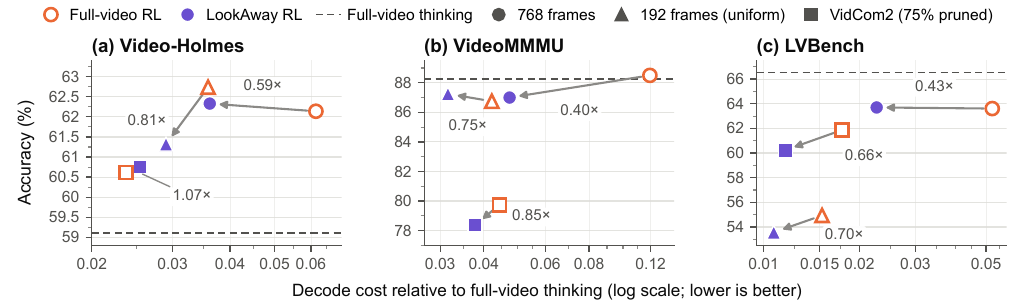}
\caption{\textbf{Less video and \method} on the held-out splits (two seeds). VidCom2~\cite{vidcom2} prunes 75\% of
the video tokens. Arrows point from full-video RL to \method RL on the same input.
Labels give the cost of \method RL relative to full-video RL on the same input.
The x-axis shows decode cost relative to full-video thinking on all 768 frames.}
\label{fig:input}
\end{figure}

\paragraph{Against less video.}
\Cref{fig:input} gives both trained models 192 of the 768 frames or 75\% VidCom2 pruning~\cite{vidcom2} (\cref{app:input}).
The savings compose.
Input reduction shrinks the video cache, while \method reduces the steps that read what remains.
On 192 frames, \method RL decodes at \N{in.lvb.cmp.rlnotes.f192-rlvan.f192.t}--\N{in.vh.cmp.rlnotes.f192-rlvan.f192.t}\x{} the cost of full-video RL on the same frames at matched accuracy, giving the lowest cost on VideoMMMU.

While input reduction alone matches \method RL on individual benchmarks, which reduction preserves accuracy depends on the videos.
Frame subsampling costs full-video RL \N{in.lvb.rlvan.f192.own.abs} points on LVBench, and token pruning costs it \N{in.mmmu.rlvan.vc2.own.abs} on VideoMMMU.
By contrast, \method RL on the full input stays within \N{in.rlnotes.maxloss} points of full-video RL on all three benchmarks.
\method writes notes while all frames remain visible, whereas an omitted frame cannot inform later reasoning.
On LVBench, its lead is largest when subsampling misses the annotated evidence (\cref{app:example}).

\FloatBarrier
\section{Conclusion}
\label{sec:conclusion}

On long videos, a VLM need not keep looking throughout its reasoning.
Across both hybrid and pure-attention VLMs, visual dependence is concentrated early in the response.
\method records timestamped notes before hiding the video's KV cache.
Without training, it nearly matches full-video accuracy on Qwen3.6-27B at low decode cost while outperforming matched-step free reasoning.
Post-note answers rely primarily on the notes, while keeping clips visible adds no significant benefit.
Jointly training note-taking and withdrawal timing with read-cost RL compresses visual access to a median of \N{e45.vh.rlnotes.hidden}--\N{e45.mmmu.rlnotes.hidden} tokens.
Across three held-out benchmarks, it writes more tokens than an identically trained full-video model yet reads only \N{test.mmmu.reads.rlnotes.rlvan.pct}--\N{test.vh.reads.rlnotes.rlvan.pct}\% as many KV positions at matched accuracy.
The savings come from cheaper steps, not shorter reasoning.

\section*{LLM Usage Statement}

We used Claude Code and Codex to edit and refine author-written manuscript text and implementation code.
For the manuscript, this assistance focused on improving language, clarity, and organization.
For the implementation, it was limited to editing and refining author-written code.

These tools were not used for literature discovery or citation selection.
They were not used for experimental design, execution, data analysis, or figure creation.
They were also not used for synthetic data generation, dataset annotation, research ideation, or scientific conclusions.
This statement concerns manuscript and code preparation rather than the VLMs evaluated in our study.
The authors take full responsibility for the final manuscript, code, results, and conclusions.

\clearpage
\pdfbookmark[1]{References}{references}
\bibliographystyle{plainnat}
\bibliography{references}

\clearpage
\pdfbookmark[0]{Appendix}{appendix}  
\beginappendix
\crefalias{section}{appendix}
\suppressfloats[t]  
\section{Limitations}
\label{app:limits}

Our RL experiments train Qwen3.6-27B with one run per protocol, while our zero-shot timing analyses also evaluate the pure-attention Qwen3-VL-32B-Thinking (\cref{app:qvl}).
In \method, hiding the video removes its KV blocks from the full-attention layers while preserving the recurrent state and video-conditioned prompt entries (\cref{app:e2}).
Because our benchmarks use multiple-choice questions, open-ended tasks requiring fine spatial detail may benefit from retaining selected clips alongside the notes.
In addition, our cost metric models decode time while keeping hidden KV blocks allocated to preserve prefix caching across requests.
\method therefore reduces memory-bandwidth reads rather than peak KV memory.
Finally, our RL training uses 2--3-minute videos, making training on hour-long videos a natural extension for long-horizon benchmarks such as LVBench.

\section{Cost Model}
\label{app:cost}

\Cref{eq:cost} uses the bf16 peak compute $F=2.25$\,PFLOP/s and memory bandwidth $W=8$\,TB/s of a single B200 GPU.
For Qwen3.6-27B, $b_{\mathrm{kv}}$ stores the keys and values across 16 full-attention layers with 4 KV heads of dimension 256, and $s_{\mathrm{loc}}=78.45$\,MB.
At each decoding step, model computation takes about 60\,$\mu$s and reading the recurrent state takes 14\,$\mu$s, whereas reading 45K video KV positions takes about 530\,$\mu$s.
Cost ratios sum over all question-seed pairs in a split.
Evaluation counts KV reads in 32-token blocks to match the attention kernel.

\paragraph{Batch size.}
\Cref{eq:cost} follows the roofline model for batched decoding.
Each step's weight read ($2N$ bytes, or \N{serve.wstep.ms}\,ms) is amortized across the $B$ concurrent requests on a model replica.
This weight-read time falls below the compute term once $B\ge\N{serve.bmin}$.
A single DGX B200 node serving one replica across its 8 GPUs ($8\times 180$\,GB) supports this batch size on our held-out prompts (\N{serve.lvb.node}--\N{serve.vh.node} concurrent requests).
Sharding the replica across the GPUs scales all terms proportionally.
At smaller batch sizes, reading the weights adds \N{serve.wstep.ms}$/B$\,ms per request at each step.
Across batch sizes from $B=160$ down to $B=32$, zero-shot \method decodes at \N{serve.vh.la0.b160}--\N{serve.mmmu.la0.b32}\x{} the held-out cost of full-video thinking (\cref{tab:serving}).

\begin{table}[t]
\centering
\small
\setlength{\tabcolsep}{3pt}
\begin{tabular}{@{}lrrr@{}}
\toprule
Setting & Video-Holmes & VideoMMMU & LVBench \\
\midrule
Eq.~(1) & 0.12 & 0.13 & 0.15 \\
$B{=}160$ & 0.14 & 0.15 & 0.17 \\
$B{=}64$ & 0.17 & 0.18 & 0.18 \\
$B{=}32$ & 0.21 & 0.22 & 0.21 \\
\bottomrule
\end{tabular}

\caption{\textbf{Decode cost of zero-shot \method by batch size}, relative to full-video thinking (held out, two
seeds). $B$ denotes concurrent requests per model replica. Eq.~(1) represents the large-$B$ limit.}
\label{tab:serving}
\end{table}

\section{Protocol and Engine Details}
\label{app:engine}

\paragraph{Prompts.}
Both prompts present the video clips in chronological order, each preceded by a timestamp label such as ``Clip 3 (00:16--00:24)''.
The clips are followed by the question, the answer options, and the task instructions.
The full-video instructions ask the model to reason through the relevant parts of the video step by step.
In \method, the following instruction replaces that sentence.
\begin{quote}\small
\emph{First watch the video and call the \texttt{put\_away\_video} tool once, passing as \texttt{notes} only the
observations from the video that you need to answer the Question, one per line: a timestamp (mm:ss), a colon, and
what is shown or said there (the people, objects, actions, on-screen text, numbers or formulas). The call puts the
video away: after it, the video is hidden from you, so work out the answer using only your notes and the
Question. Call the tool exactly once. After the call and your remaining thinking, reply with the
\texttt{<answer>} block alone.}
\end{quote}
Placed in the system turn, the tool schema describes \texttt{put\_away\_video} as hiding the video once the model has recorded the needed observations.
It also states that ``the instructions, the Question, your notes and everything you have written stay visible''.
Each response begins with forced tokens that close an empty thinking block and open the tool call.
Once the model closes the call, a forced tool response states ``The video is now hidden. Continue from your notes and the Question.''
A new thinking block then opens for subsequent reasoning.

\paragraph{Decoding.}
Thinking mode samples with temperature 1.0, top-$p$ 0.95, and top-$k$ 20.
Non-thinking mode uses the model card defaults (temperature 0.7, top-$p$ 0.8, top-$k$ 20, and presence penalty 1.5).
Responses are capped at 32{,}768 output tokens or the remaining context window.
A repetition guard forces the final answer if a 64-token window repeats eight times or the response approaches the length cap.
We parse the answer letter from the final \texttt{<answer>} block after the thinking block, counting unparseable outputs as incorrect.

\paragraph{Hiding.}
Rounding retained spans outward to 32-token blocks (\cref{sec:engine}) leaves at most 31 video positions readable at each boundary.
Clip timestamp labels are hidden along with the video tokens.
Because prefill is unmasked and the KV cache is not modified in place, prefix caching across requests remains intact.
All forced tokens (opening the call, the tool response, and pre-filled control calls) are counted in decoding steps and KV reads.
While forced tokens open the tool call, the model itself chooses when to close it.
Zero-shot, the call remains unclosed in at most \N{eff.lvb.notes.open}\% of development responses and \N{serve.open.la0.max}\% of held-out responses.
After RL training, unclosed calls occur in at most \N{serve.open.larl.max}\% of responses, and all such responses are billed at their full read cost.
During evaluation, the plugin's online read counts match an offline protocol replay on every request.
During training, any rollout whose online count differs from the offline replay is discarded.

\paragraph{Controls.}
Withdraw@$\tau$ hides the video at the seed-averaged token position where \method closes its call on the same question.
Both seeds of a question therefore withdraw the video at the exact same step.
In Withdraw@start, the forced empty call spans 28 tokens decoded with the video visible.
Notes + video samples fresh notes with the video kept visible throughout.
For clip-retention controls, selected clips are retained in full.
Model-selected clips take the first four clips named in the call, whereas SigLIP selects the four clips with the highest frame-to-question similarity (excluding answer options).
Forced own notes replay the notes from seed 0.
Shuffling permutes note lines, falling back to sentences or words when the notes contain only one line.
Budget@$N$ appends the end-of-thinking delimiter after $N$ thinking tokens~\cite{s1}, after which the model generates its reply normally.

\paragraph{Non-thinking mode.}
In non-thinking mode, \method uses the same tool schema, tool response, and engine state machine.
The final instruction sentence simply changes to ``After the call, reply with the \texttt{<answer>} block.''
The tool call opens at the start of the reply and closes after a median of \N{eff.mmmu.ntnotes.hidden}, \N{eff.vh.ntnotes.hidden}, and \N{eff.lvb.ntnotes.hidden} tokens on VideoMMMU-dev, Video-Holmes-dev, and LVBench-dev.
Because non-thinking replies are brief, the note-taking phase accounts for a larger fraction of total tokens on short responses.
When splitting Video-Holmes-dev at the median reply length (\N{eff.vh.split.cut} tokens), \method decodes at \N{eff.vh.split.short}\x{} the full-video cost on the shorter half and \N{eff.vh.split.long}\x{} on the longer half.

\section{Training Details}
\label{app:training}

\paragraph{Data.}
We retain a training question only if the base model fails to answer it without the video (\cref{sec:setup}).
We keep at most two questions per video and partition the videos into supervised, RL, and validation splits.
After filtering out prompts longer than 46{,}567 tokens, the RL set contains 542 questions drawn from the supervised and RL partitions (470 from LLaVA-Video and 72 from Video-Holmes).

\paragraph{Recipe.}
Supervised fine-tuning runs for one epoch with a batch size of 32, a learning rate of $10^{-6}$, and a 5\% warmup.
RL training uses GRPO with group size $G=8$ and 16 prompts per step across two mini-batches.
We use asymmetric clipping $[0.2, 0.28]$, a token-mean loss, no KL or entropy penalty, and unnormalized advantages (without dividing by the group standard deviation).
Training runs for 150 steps at a constant learning rate of $10^{-6}$.
Rollouts sample at temperature 1.0 and top-$p$ 1.0 with a maximum of 8{,}192 output tokens and the same repetition guard.
Forced tokens are masked out of the policy loss.
We freeze the vision encoder and the Gated DeltaNet decay parameters, and both RL models share the exact same random seed, data order, and hyperparameters.

\paragraph{Per-token read cost.}
In \cref{eq:reward,eq:adv}, $a_{i,s}$ counts KV positions per token rather than per 32-token block.
The prompt term in $A_i$ follows the prompt-charged read accounting of \cite{da}, which treats video prefill as cached.
The reference read count $a_{\mathrm{ref}}$ is the mean of $A$ across the base model's two full-video rollouts on the same question.
When computing the group mean $\bar c_t$, we set $c_{j,t}=0$ for steps $t$ beyond the end of response $j$.
We implement the centered cost term via the equivalent leave-one-out identity $c_{i,t}-\bar c_t=\frac{G-1}{G}\big(c_{i,t}-\frac{1}{G-1}\sum_{j\neq i}c_{j,t}\big)$.
Whenever the free budget $f\,a_{\mathrm{ref}}$ exceeds the sum of the prompt term and the first step's reads, the first token's cost-to-go equals $E_i$, so $\hat A_{i,1}$ recovers the standard GRPO advantage of $R_i$.

\section{Intervals for the Development Analyses}
\label{app:intervals}

\begin{table}[t]
\centering
\small
\setlength{\tabcolsep}{3pt}
\begin{tabular}{@{}l ll@{}}
\toprule
Difference (points) & VideoMMMU-dev & Video-Holmes-dev \\
\midrule
\multicolumn{3}{@{}l}{\emph{Timing of visual access} (\cref{sec:timing})} \\
\method $-$ Withdraw@$\tau$ & \N{e1.mmmu.primary} \N{e1.mmmu.primary.ci} & \N{e1.vh.primary} \N{e1.vh.primary.ci} \\
Withdraw@512 $-$ full video & \N{x.mmmu.fixed-512-vanilla} \N{x.mmmu.fixed-512-vanilla.ci} & \N{x.vh.fixed-512-vanilla} \N{x.vh.fixed-512-vanilla.ci} \\
Notes + video $-$ \method & \N{e1.mmmu.notes-visible-notes} \N{e1.mmmu.notes-visible-notes.ci} & \N{e1.vh.notes-visible-notes} \N{e1.vh.notes-visible-notes.ci} \\
Withdraw@$\tau$ $-$ full video & \N{x.mmmu.position-matched-vanilla} \N{x.mmmu.position-matched-vanilla.ci} & \N{x.vh.position-matched-vanilla} \N{x.vh.position-matched-vanilla.ci} \\
Withdraw@start $-$ full video & \N{x.mmmu.blind-vanilla} \N{x.mmmu.blind-vanilla.ci} & \N{x.vh.blind-vanilla} \N{x.vh.blind-vanilla.ci} \\
Clips only $-$ full video & \N{x.mmmu.pinsonly-vanilla} \N{x.mmmu.pinsonly-vanilla.ci} & \N{x.vh.pinsonly-vanilla} \N{x.vh.pinsonly-vanilla.ci} \\
Notes + video $-$ full video & \N{x.mmmu.notes-visible-vanilla} \N{x.mmmu.notes-visible-vanilla.ci} & \N{x.vh.notes-visible-vanilla} \N{x.vh.notes-visible-vanilla.ci} \\
Withdraw@256 $-$ full video & \N{bd.mmmu.wd256.d} \N{bd.mmmu.wd256.d.ci} & \N{bd.vh.wd256.d} \N{bd.vh.wd256.d.ci} \\
Withdraw@128 $-$ full video & \N{bd.mmmu.wd128.d} \N{bd.mmmu.wd128.d.ci} & \N{bd.vh.wd128.d} \N{bd.vh.wd128.d.ci} \\
Withdraw@64 $-$ full video & \N{bd.mmmu.wd64.d} \N{bd.mmmu.wd64.d.ci} & \N{bd.vh.wd64.d} \N{bd.vh.wd64.d.ci} \\
Budget@1024 $-$ full video & \N{bd.mmmu.bf1024.d} \N{bd.mmmu.bf1024.d.ci} & \N{bd.vh.bf1024.d} \N{bd.vh.bf1024.d.ci} \\
Budget@512 $-$ full video & \N{bd.mmmu.bf512.d} \N{bd.mmmu.bf512.d.ci} & \N{bd.vh.bf512.d} \N{bd.vh.bf512.d.ci} \\
Budget@256 $-$ full video & \N{bd.mmmu.bf256.d} \N{bd.mmmu.bf256.d.ci} & \N{bd.vh.bf256.d} \N{bd.vh.bf256.d.ci} \\
\method $-$ full video & \N{x.mmmu.notes-vanilla} \N{x.mmmu.notes-vanilla.ci} & \N{x.vh.notes-vanilla} \N{x.vh.notes-vanilla.ci} \\
Non-thinking, \method $-$ full video & \N{eff.mmmu.ntnotes.d} \N{eff.mmmu.ntnotes.d.ci} & \N{eff.vh.ntnotes.d} \N{eff.vh.ntnotes.d.ci} \\
\midrule
\multicolumn{3}{@{}l}{\emph{Notes versus pixels} (\cref{sec:pixels})} \\
Own notes replayed $-$ free generation & \N{e45.mmmu.own-notes} \N{e45.mmmu.own-notes.ci} & \N{e45.vh.own-notes} \N{e45.vh.own-notes.ci} \\
Own notes $-$ another video's notes & \N{e45.mmmu.own-swap} \N{e45.mmmu.own-swap.ci} & \N{e45.vh.own-swap} \N{e45.vh.own-swap.ci} \\
Another video's notes $-$ Withdraw@start & \N{e45.mmmu.swap-blind} \N{e45.mmmu.swap-blind.ci} & \N{e45.vh.swap-blind} \N{e45.vh.swap-blind.ci} \\
Own notes $-$ shuffled notes & \N{e45.mmmu.own-shuffle} \N{e45.mmmu.own-shuffle.ci} & \N{e45.vh.own-shuffle} \N{e45.vh.own-shuffle.ci} \\
Notes + clips (model's) $-$ \method & \N{gate.mmmu.pins-notes} \N{gate.mmmu.pins-notes.ci} & \N{gate.vh.pins-notes} \N{gate.vh.pins-notes.ci} \\
Model's clips $-$ SigLIP clips & \N{gate.mmmu.pins-pinssiglip} \N{gate.mmmu.pins-pinssiglip.ci} & \N{gate.vh.pins-pinssiglip} \N{gate.vh.pins-pinssiglip.ci} \\
Model's clips $-$ random clips & \N{gate.mmmu.pins-pinsrandom} \N{gate.mmmu.pins-pinsrandom.ci} & \N{gate.vh.pins-pinsrandom} \N{gate.vh.pins-pinsrandom.ci} \\
After RL, notes and clips $-$ notes & \N{e45.mmmu.RLpins-RLnotes} \N{e45.mmmu.RLpins-RLnotes.ci} & \N{e45.vh.RLpins-RLnotes} \N{e45.vh.RLpins-RLnotes.ci} \\
\bottomrule
\end{tabular}
\caption{\textbf{Development contrasts with 95\% intervals} (10{,}000 bootstrap resamples). Results use Qwen3.6-27B across two seeds and are zero-shot except for the After-RL row. The Notes + clips and Model's clips rows come from a separate zero-shot evaluation.}
\label{tab:contrasts}
\end{table}

\Cref{tab:contrasts} reports the 95\% bootstrap confidence intervals for all development contrasts in \cref{sec:timing,sec:pixels}.

\section{A Pure-Attention Model}
\label{app:qvl}

\begin{table}[t]
\centering
\small
\begin{tabular}{@{}l rrrrr rrrrr@{}}
\toprule
 & \multicolumn{5}{c}{VideoMMMU-dev} & \multicolumn{5}{c}{Video-Holmes-dev} \\
\cmidrule(lr){2-6}\cmidrule(l){7-11}
Condition & Acc & $\Delta$ & Tokens & Cost & $\tau$ & Acc & $\Delta$ & Tokens & Cost & $\tau$ \\
\midrule
Full video & 79.2 & -- & 1,359 & 1 & -- & 67.0 & -- & 709 & 1 & -- \\
Withdraw@$\tau$ & 81.0 & +1.8 & 1,458 & 0.286 & 240 & 66.9 & $-$0.2 & 777 & 0.550 & 306 \\
Withdraw@start & 69.0 & $-$10.3 & 1,495 & 0.068 & 22 & 63.2 & $-$3.9 & 1,110 & 0.143 & 22 \\
\method & 79.0 & $-$0.2 & 1,776 & 0.362 & 238 & 62.6 & $-$4.4 & 1,679 & 1.309 & 342 \\
\midrule
Difference (points) & \multicolumn{5}{c}{$\Delta$ [95\%]} & \multicolumn{5}{c}{$\Delta$ [95\%]} \\
\midrule
Withdraw@start $-$ full video & \multicolumn{5}{c}{$-$10.3 [$-$15.5, $-$5.0]} & \multicolumn{5}{c}{$-$3.9 [$-$6.5, $-$1.3]} \\
Withdraw@$\tau$ $-$ full video & \multicolumn{5}{c}{+1.8 [$-$1.5, +5.0]} & \multicolumn{5}{c}{$-$0.2 [$-$1.6, +1.4]} \\
\bottomrule
\end{tabular}

\caption{\textbf{Zero-shot Qwen3-VL-32B-Thinking} across two seeds. Columns follow \cref{tab:access}, and $\tau$ denotes the median response tokens before video withdrawal. Withdraw@$\tau$ responses that finish before $\tau$ (\N{qvl.mmmu.posmatch.kept}\% and \N{qvl.vh.posmatch.kept}\%) retain the video throughout. The bottom rows report accuracy differences with 95\% intervals.}
\label{tab:qvl}
\end{table}

To test whether visual front-loading depends on a recurrent state, we evaluate the pure-attention Qwen3-VL-32B-Thinking~\cite{qwen3vl}, where all 64 layers use full attention.
Without a recurrent state, no visual information bypasses attention once the video KV blocks are hidden.
Prompts, clips, controls, and statistics follow \cref{tab:access}.
We use the model's native tool format, model-card sampling settings, and hardware constants (four times the KV bytes per token of Qwen3.6-27B and no recurrent state).
Withdraw@$\tau$ uses this model's own \method positions.
Visual access is front-loaded on this model as well (\cref{tab:qvl}, bottom).
On VideoMMMU-dev, zero-shot \method matches full-video accuracy at 0.36\x{} the decode cost.
On Video-Holmes-dev, the base model thinks briefly on the full video (\N{qvl.vh.full.tok} tokens, versus \N{e1.vh.vanilla.tok} for Qwen3.6-27B), whereas zero-shot notes lengthen the response to \N{qvl.vh.notes.tok} tokens (\N{qvl.mmmu.notes.tok} on VideoMMMU-dev).
This contrast motivates read-cost RL (\cref{sec:train}), which we apply to Qwen3.6-27B.

\section{Notes Without the Video}
\label{app:e2}

The note interventions in \cref{sec:pixels} encode the video during prefill and hide it after the call.
The Gated DeltaNet recurrent state and the KV entries of the question and notes are still computed with the video in view.
To separate these channels from the note text, we pass the model's own notes to a prompt without the video.
The clip section is omitted while all other tokens, including the forced call and tool reply, stay unchanged.
A second condition forces an empty call on the same prompt, leaving the question alone.
Both use the development splits, two seeds, and the settings of the note interventions.

\begin{table}[t]
\centering
\small
\setlength{\tabcolsep}{3pt}
\begin{tabular}{@{}l rrr rrr@{}}
\toprule
 & \multicolumn{3}{c}{VideoMMMU-dev} & \multicolumn{3}{c}{Video-Holmes-dev} \\
\cmidrule(lr){2-4}\cmidrule(l){5-7}
Condition & Acc & Tokens & Cost & Acc & Tokens & Cost \\
\midrule
Own notes, video in prefill & 83.0 & 2,043 & 0.125 & 70.6 & 1,231 & 0.121 \\
Own notes, no video & 79.8 & 1,887 & 0.067 & 66.1 & 1,547 & 0.082 \\
Withdraw@start, video in prefill & 69.0 & 4,073 & 0.179 & 65.3 & 1,858 & 0.133 \\
No notes, no video & 57.8 & 3,893 & 0.144 & 46.8 & 3,104 & 0.180 \\
\method{}, video in prefill & 83.2 & 2,025 & 0.133 & 70.9 & 1,199 & 0.118 \\
\midrule
Difference (points) & \multicolumn{3}{c}{$\Delta$ [95\%]} & \multicolumn{3}{c}{$\Delta$ [95\%]} \\
\midrule
Own notes (no video $-$ video in prefill) & \multicolumn{3}{c}{$-$3.2 [$-$6.8, 0.0]} & \multicolumn{3}{c}{$-$4.5 [$-$6.4, $-$2.5]} \\
No notes (video in prefill $-$ no video) & \multicolumn{3}{c}{+11.2 [+5.8, +16.8]} & \multicolumn{3}{c}{+18.5 [+14.8, +22.0]} \\
No video (own notes $-$ no notes) & \multicolumn{3}{c}{+22.0 [+15.0, +29.0]} & \multicolumn{3}{c}{+19.3 [+15.5, +23.0]} \\
Own notes, no video $-$ \method{} & \multicolumn{3}{c}{$-$3.5 [$-$6.8, $-$0.5]} & \multicolumn{3}{c}{$-$4.8 [$-$7.1, $-$2.3]} \\
\bottomrule
\end{tabular}

\caption{\textbf{Notes without the video} (zero-shot, two seeds).
In the video-in-prefill condition, the video is encoded during prefill and hidden after the call.
In the no-video condition, the prompt never includes the video. Cost is relative to full-video thinking.}
\label{tab:e2}
\end{table}

In the no-video prompt, the written notes alone gain 22.0 points over the question alone on VideoMMMU-dev and 19.3 points on Video-Holmes-dev (\cref{tab:e2}).
Removing the prefill channels reduces accuracy by \N{e2.vh.textown-own.abs} points on Video-Holmes-dev (interval below zero) and \N{e2.mmmu.textown-own.abs} points on VideoMMMU-dev (interval reaching zero).
When nothing is written, those prefill channels alone contribute \N{e2.mmmu.blind-textempty} and \N{e2.vh.blind-textempty} points (Withdraw@start versus the question alone).
Withdraw@start is thus above a no-video floor, so its accuracy drop in \cref{sec:timing} is a conservative estimate of visual contribution.
\method keeps both the written notes and these prefill channels because it hides the video only during decoding (last row of \cref{tab:e2}).

\section{Why the Video Can Go}
\label{app:mech}

\begin{figure}[t]
\begin{sidecaption}{0.5}{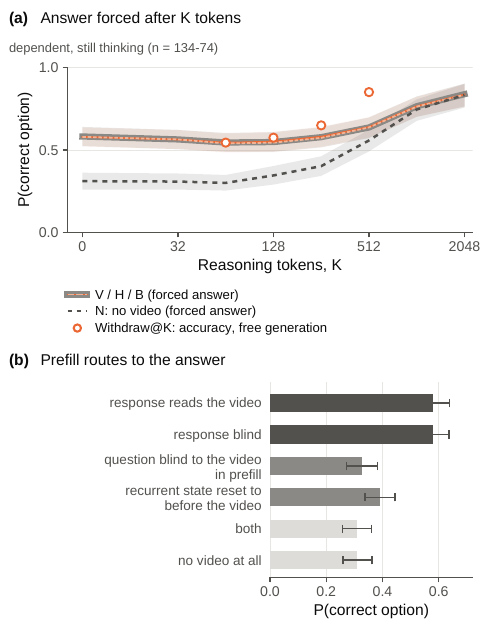}
\caption{\textbf{The answer follows the text and the question} (dependent questions).
(a) Probability of the correct option when the answer is forced after $K$ tokens of a full-video response.
Conditions keep reading the video (V), hide it after $K$ tokens (H), hide it from the start of decoding (B), or omit it from the prompt (N).
Only responses still thinking at $K$ are included, and hollow markers show free-generation accuracy for Withdraw@$K$.
(b) Correct-option probability when the answer is forced right after prefill under selective attention masks, with 95\% intervals.}
\label{fig:mech-probe}
\end{sidecaption}
\end{figure}

\paragraph{Method.}
We re-score saved zero-shot responses under their own prompts while varying only what the model may read.
The sample has \N{mech.n} questions from the two development splits.
It includes all \N{mech.ndep} dependent questions where full-video thinking is right and Withdraw@start is wrong in seed 0, plus 65 random questions.
For each question, we take the first 8{,}192 tokens of the seed-0 responses from full-video thinking, \method, and Notes + video.
Exact attention weights averaged over all heads and all 16 full-attention layers provide the attention statistics.
Hiding follows the engine (\cref{sec:engine}), removing clip KV blocks from the full-attention layers while leaving the recurrent state intact.
We compare four regimes during teacher-forced scoring.
Condition V reads the video throughout, Condition H hides it after $K$ tokens, Condition B hides it from the start, and Condition N omits the clips from the prompt.
To read the answer after $K$ tokens, we append the close of the thinking block and \texttt{<answer} tag and compute the renormalized probability of the correct option.
On seed-1 responses outside the sample, the engine's log-probabilities match a reference forward pass within \N{mech.valid.visible} nats per token.
With the video hidden, they match the training attention mask within \N{mech.valid.masked} nats, within the \N{mech.valid.floor}-nat kernel baseline gap.
Hiding the video shifts log-probabilities by \N{mech.valid.control} nats.

\paragraph{The notes read the frames they cite.}
We analyze \N{mech.p4.lines} timestamped note lines with at least three tokens after the timestamp (at most eight per response).
As each line is written, attention concentrates on the clip covering its cited time and neighboring clips.
Attention to these clips reaches \N{mech.p4.ecited}\x{} \N{mech.p4.ecited.ci} their share of the video's keys, versus \N{mech.p4.eother}\x{} \N{mech.p4.eother.ci} for times cited by other lines.
Hiding only the cited clips lowers the log-probability of these words by \N{mech.p4.kocited} nats per token, whereas hiding as many random clips lowers it by \N{mech.p4.korandom} nats (paired difference \N{mech.p4.b} \N{mech.p4.b.ci}).

\paragraph{After the notes, attention leaves the video.}
In Notes + video, the video stays readable throughout.
We re-score its responses with the tool reply replaced by ``Notes received.'' so no reply says the video is hidden.
Comparing the notes with the 482 tokens after the reply, attention to the video relative to its key share falls by \N{mech.p5.drop}\% \N{mech.p5.drop.ci}.
This drop occurs in \N{mech.p5.share}\% of the questions, while attention to the question and instructions holds (ratio \N{mech.p5.qratio} \N{mech.p5.qratio.ci}).
At matched steps of full-video thinking on the same questions, video attention falls by \N{mech.p5.fulldrop}\% \N{mech.p5.fulldrop.ci}.
The notes thus accelerate visual disengagement beyond position alone (difference in differences \N{mech.p5.did} \N{mech.p5.did.ci}).
Tokens after the notes also depend little on the video.
The KL between their next-token distributions with the video readable versus hidden is \N{mech.p3.nv} nats per token, against \N{mech.p3.van} for full-video thinking at matched steps (ratio \N{mech.p3.ratio} \N{mech.p3.ratio.ci}).

\paragraph{The answer follows the text and the question.}
Forcing the answer after $K$ tokens of a full-video response yields the same correct-option probability whether the video stays readable or is hidden at $K$ (\cref{fig:mech-probe}a).
On the dependent questions, V and H differ by \N{mech.p1.gap0} \N{mech.p1.gap0.ci} points at $K=0$ and \N{mech.p1.gap512} \N{mech.p1.gap512.ci} at $K=512$, and B tracks both.
By contrast, free generation under Withdraw@$K$ (hollow markers) loses accuracy when the video disappears early.
Thus, what the video adds after $K$ steps enters the answer through the text the model writes.
Without any video (N), correct-option probability starts at \N{mech.n.c0} (versus \N{mech.v.c0}) and reaches \N{mech.n.c1024} (versus \N{mech.v.c1024}) after 1{,}024 tokens across the \N{mech.n.c1024.count} responses still thinking there.
Before any text is written, the video reaches the answer through the question (\cref{fig:mech-probe}b).
Blocking the question and instructions from attending to the clips during prefill drops the immediate answer from \N{mech.s1.f} to \N{mech.s1.qb}.
This matches the no-video level of \N{mech.s1.n} (difference \N{mech.s1.qbn} \N{mech.s1.qbn.ci} points) even though the recurrent state has read the clips.
When they attend to the clips in full-attention layers but receive the pre-video recurrent state, probability drops only to \N{mech.s1.gs}.
\method keeps both channels because it hides the video only during decoding (\cref{app:e2}).

\section{Full Held-out Results}
\label{app:full}

\begin{table}[t]
\centering
\small
\setlength{\tabcolsep}{4pt}
\resizebox{\textwidth}{!}{\begin{tabular}{@{}l ll ll ll@{}}
\toprule
 & \multicolumn{2}{c}{Video-Holmes} & \multicolumn{2}{c}{VideoMMMU} & \multicolumn{2}{c}{LVBench} \\
\cmidrule(lr){2-3}\cmidrule(lr){4-5}\cmidrule(l){6-7}
Model & $\Delta$ [95\%] & Cost [95\%] & $\Delta$ [95\%] & Cost [95\%] & $\Delta$ [95\%] & Cost [95\%] \\
\midrule
Full video, non-thinking & $-$0.5 [$-$2.1, +1.1] & 0.102 [0.097, 0.107] & $-$1.5 [$-$3.5, +0.5] & 0.200 [0.178, 0.223] & $-$0.3 [$-$2.3, +1.7] & 0.060 [0.053, 0.068] \\
\method{}, zero-shot & $-$1.6 [$-$3.0, $-$0.2] & 0.119 [0.114, 0.125] & $-$1.5 [$-$3.5, +0.5] & 0.127 [0.114, 0.141] & $-$1.0 [$-$3.0, +1.0] & 0.155 [0.121, 0.195] \\
Full-video SFT & $-$0.2 [$-$1.6, +1.2] & 0.928 [0.897, 0.961] & $-$1.0 [$-$2.6, +0.5] & 0.952 [0.895, 1.016] & $-$1.1 [$-$2.8, +0.5] & 0.952 [0.887, 1.020] \\
\method{} SFT & $-$1.4 [$-$2.8, 0.0] & 0.132 [0.126, 0.138] & $-$0.9 [$-$2.8, +1.0] & 0.128 [0.115, 0.142] & $-$2.1 [$-$4.2, $-$0.1] & 0.156 [0.119, 0.201] \\
Full-video RL & +3.0 [+1.3, +4.7] & 0.062 [0.059, 0.064] & +0.2 [$-$1.8, +2.2] & 0.119 [0.104, 0.138] & $-$2.9 [$-$5.2, $-$0.8] & 0.052 [0.047, 0.059] \\
\method{} RL & +3.2 [+1.5, +4.9] & 0.036 [0.035, 0.038] & $-$1.2 [$-$3.5, +1.1] & 0.047 [0.039, 0.058] & $-$2.8 [$-$4.9, $-$0.7] & 0.023 [0.020, 0.026] \\
\midrule
\method{} RL vs.\ full-video RL & +0.2 [$-$1.3, +1.7] & 0.587 [0.572, 0.604] & $-$1.5 [$-$3.6, +0.5] & 0.397 [0.342, 0.460] & +0.1 [$-$1.9, +2.1] & 0.433 [0.422, 0.444] \\
\bottomrule
\end{tabular}
}
\caption{\textbf{Held-out results with 95\% intervals} (Qwen3.6-27B, two seeds). $\Delta$ (points) and Cost are
relative to untrained full-video thinking unless the row specifies another reference.}
\label{tab:full}
\end{table}

\Cref{tab:full} reports the complete held-out comparisons from \cref{tab:test} along with their 95\% bootstrap confidence intervals.

\section{Less Video}
\label{app:input}

\begin{table}[t]
\centering
\scriptsize
\setlength{\tabcolsep}{3pt}
\resizebox{\textwidth}{!}{\begin{tabular}{@{}l ll ll ll@{}}
\toprule
 & \multicolumn{2}{c}{Video-Holmes} & \multicolumn{2}{c}{VideoMMMU} & \multicolumn{2}{c}{LVBench} \\
\cmidrule(lr){2-3}\cmidrule(lr){4-5}\cmidrule(l){6-7}
Row / comparison & $\Delta$ [95\%] & Cost [95\%] & $\Delta$ [95\%] & Cost [95\%] & $\Delta$ [95\%] & Cost [95\%] \\
\midrule
Full-video RL, 192 frames & +3.6 [+1.9, +5.4] & 0.036 [0.034, 0.038] & $-$1.5 [$-$3.6, +0.6] & 0.042 [0.037, 0.049] & $-$11.6 [$-$14.9, $-$8.4] & 0.015 [0.014, 0.017] \\
Full-video RL, VidCom2 & +1.5 [$-$0.3, +3.4] & 0.024 [0.023, 0.025] & $-$8.5 [$-$11.9, $-$5.4] & 0.044 [0.038, 0.052] & $-$4.7 [$-$7.1, $-$2.3] & 0.018 [0.016, 0.020] \\
\method{} RL, 192 frames & +2.2 [+0.4, +4.0] & 0.029 [0.028, 0.031] & $-$1.0 [$-$3.2, +1.2] & 0.032 [0.026, 0.038] & $-$13.0 [$-$16.3, $-$9.7] & 0.011 [0.010, 0.012] \\
\method{} RL, VidCom2 & +1.6 [$-$0.1, +3.4] & 0.025 [0.024, 0.027] & $-$9.9 [$-$13.4, $-$6.5] & 0.038 [0.030, 0.047] & $-$6.3 [$-$8.4, $-$4.2] & 0.012 [0.010, 0.013] \\
Full video, non-thinking, 192 frames & $-$1.3 [$-$2.9, +0.4] & 0.050 [0.047, 0.053] & $-$0.2 [$-$2.2, +1.8] & 0.070 [0.063, 0.077] & $-$12.3 [$-$15.5, $-$9.2] & 0.030 [0.026, 0.034] \\
Full video, non-thinking, VidCom2 & $-$3.1 [$-$4.7, $-$1.5] & 0.035 [0.034, 0.037] & $-$9.8 [$-$12.8, $-$6.8] & 0.095 [0.083, 0.109] & $-$5.2 [$-$7.5, $-$2.9] & 0.024 [0.021, 0.027] \\
\midrule
Full-video RL, 192 vs.\ \method{} RL & +0.4 [$-$1.2, +2.1]$^\dagger$ & 0.992 [0.969, 1.015] & $-$0.2 [$-$2.5, +1.9]$^\dagger$ & 0.890 [0.769, 1.028] & $-$8.8 [$-$11.5, $-$6.0] & 0.674 [0.656, 0.692] \\
Full-video RL, VidCom2 vs.\ \method{} RL & $-$1.7 [$-$3.5, +0.1] & 0.657 [0.643, 0.671] & $-$7.2 [$-$10.4, $-$4.2] & 0.936 [0.785, 1.112] & $-$1.8 [$-$4.1, +0.3] & 0.779 [0.760, 0.799] \\
Same input (192), \method{} vs.\ full-video RL & $-$1.4 [$-$3.0, +0.2] & 0.810 [0.787, 0.834] & +0.5 [$-$1.6, +2.6] & 0.749 [0.641, 0.876] & $-$1.4 [$-$3.6, +0.9] & 0.705 [0.688, 0.722] \\
Same input (VidCom2), \method{} vs.\ full-video RL & +0.1 [$-$1.6, +1.8] & 1.073 [1.030, 1.120] & $-$1.4 [$-$3.9, +1.1] & 0.851 [0.711, 1.015] & $-$1.6 [$-$3.7, +0.4] & 0.662 [0.636, 0.696] \\
\bottomrule
\end{tabular}
}
\caption{\textbf{Reduced inputs on the held-out splits} (Qwen3.6-27B, two seeds) with 95\% intervals.
The top rows compare against full-video thinking on all 768 frames.
The bottom rows compare against the reference named after ``vs.'' (using 768 frames unless marked as same input).
$\Delta$ is in percentage points, and $^\dagger$ marks configurations within 0.5 points of the reference at equal or lower cost in point estimate.}
\label{tab:inputfull}
\end{table}

To construct the 192-frame input without re-decoding the raw video, we replace each target frame in the 192-frame schedule with the nearest frame from the 768-frame pool.
Temporal offsets are at most \N{in.vh.nnerr}, \N{in.mmmu.nnerr}, and \N{in.lvb.nnerr}\,s on Video-Holmes, VideoMMMU, and LVBench.
Clip labels therefore match the displayed frames while reducing only temporal density.
For token pruning, we apply VidCom2~\cite{vidcom2} in vLLM 0.29 at a 75\% pruning rate.
Each 16-frame clip retains one quarter of its encoded video tokens, which vLLM places at the clip's first frame pair with recomputed rotary position embeddings.
When comparing models on the same compressed input, \method's relative decode saving scales with the video's share of the KV cache.
Because per-step compute and recurrent-state reads remain fixed, shrinking the initial video cache leaves a smaller fraction of per-step cost to remove (same-input rows of \cref{tab:inputfull}).

\section{Examples}
\label{app:example}

\begin{figure}[t]
\centering
\includegraphics[width=\textwidth]{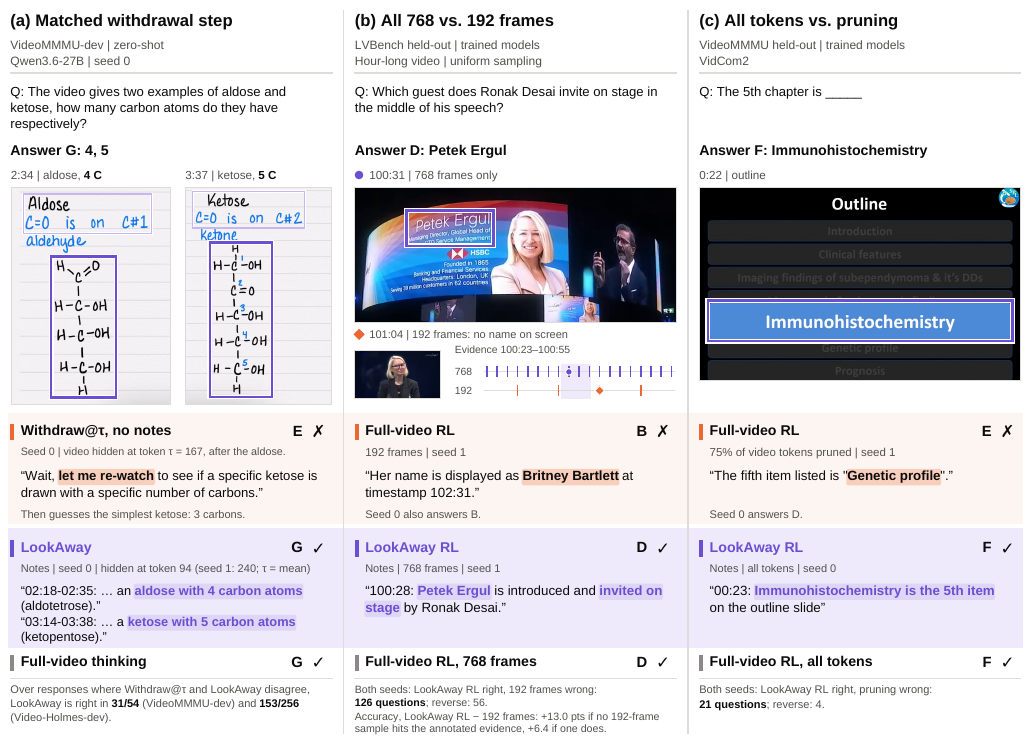}
\caption{\textbf{Qualitative comparisons.} In each panel, \method (purple) answers correctly across both seeds while the baseline condition (orange) answers incorrectly. Gray shows the full-video reference, and bounding boxes highlight the key visual evidence. Purple highlights show where \method records this evidence in its notes, while orange highlights show the baseline's reasoning. All text quotes are verbatim, and sampled frames are shown at source resolution.}
\label{fig:qual}
\end{figure}

\begin{figure}[t]
\centering
\small
\fcolorbox{black!25}{white}{\begin{minipage}{0.97\textwidth}
\begin{minipage}[t]{0.485\linewidth}
\textbf{Question} (Video-Holmes test). What is the intention of the blonde woman covering her phone to her chest
after turning it off? (A)~Relieve one's own fear (B)~Check if the phone is damaged (C)~Try to restart the phone
(D)~Avoid being discovered by the masked person (E)~Prevent the phone from continuing to take screenshots
(F)~Save power\\[3pt]
{\color{notespurple}\textbf{Notes}} (video visible)\\
\texttt{\footnotesize 0:47: the blonde woman turns off her phone and hugs it to her chest, appearing scared after seeing the ghost figure}
\end{minipage}\hfill
\begin{minipage}[t]{0.485\linewidth}
\textbf{Reasoning} (video hidden). ``She turns off her phone, likely to turn off the light or because she's
terrified, and then hugs it to her chest. This is a classic defensive posture when someone is frightened, trying
to protect themselves or calm their fear.'' \quad Answer \textbf{A} $\times$\\[3pt]
{\footnotesize 325 output tokens. The other seed notes the gesture at 0:46, ``trying to comfort herself'', and also
answers A. The explanation comes more than half a minute earlier. In the first 15\,s she sets a double tap on the
back of the phone to take a screenshot, and the film ends on a ``Screenshot'' card. Full-video thinking answers E in
both seeds (2{,}354 and 4{,}097 tokens), linking the gesture to that setting. Full-video RL answers A in both.}
\end{minipage}
\end{minipage}}
\caption{\textbf{Error analysis when notes omit an early setup.} Here the note records the gesture at 0:47 but misses the setup in the first 15\,s. Across the held-out sets, \method RL is correct and full-video thinking is incorrect in both seeds on 87 Video-Holmes and 42 LVBench questions, versus 67 and 61 in the reverse direction. On 32 of those 67 Video-Holmes questions and 28 of the 61 LVBench questions, full-video RL makes the same error in both seeds.}
\label{fig:failure}
\end{figure}

\Cref{fig:qual} illustrates three examples where \method answers correctly in both seeds while the baseline fails.
Under Withdraw@$\tau$, the model lacks notes, writes ``let me re-watch'', and guesses the carbon counts of two sugars that \method explicitly recorded.
Under 192-frame subsampling, uniform sampling misses the brief window showing a guest's name, causing full-video RL to name the wrong person while \method RL on 768 frames records the correct name.
Across LVBench, \method RL is correct and 192-frame full-video RL is incorrect in both seeds on \N{qual.b.win} questions, compared with \N{qual.b.rev} in the reverse direction (\cref{sec:test}).
Similarly, under 75\% token pruning, full-video RL misreads the fifth line of an outline slide that \method RL transcribes accurately.
Finally, \cref{fig:failure} illustrates how downstream reasoning follows the notes even when they are incomplete.
When the note captures a character's gesture at 0:47 but omits the setup in the first 15\,s, both seeds infer the wrong motive from the gesture alone (an error also shared by full-video RL).
All response quotes are verbatim with ellipses marking omitted text.

\end{document}